\documentclass[fleqn,10pt]{wlscirep}
\usepackage[utf8]{inputenc}
\usepackage[T1]{fontenc}
\usepackage{multirow}
\usepackage{tikz}

\title{BoxCell: Leveraging SAM for Cell Segmentation \\ with Box Supervision}

\author[1,*]{Aayush Kumar Tyagi}
\author[2]{Vaibhav Mishra}
\author[3]{Prathosh A.P.}
\author[1,2]{Mausam}
\affil[1]{India Institute Of Technology, Yardi School of Artificial Intelligence, New Delhi, 110016, India}
\affil[2]{India Institute Of Technology, Department of Computer Science and Engineering, New Delhi, 110016, India}
\affil[3]{Indian Institute of Science, Department of Electrical Communication Engineering, Bengaluru, 560012, India}

\affil[*]{aiz218615@scai.iitd.ac.in}

\begin{abstract}
Cell segmentation in histopathological images is vital for diagnosis, and treatment of several diseases. Annotating data is tedious, and requires medical expertise, making it difficult to employ supervised learning. Instead, we study a weakly supervised setting, where only bounding box supervision is available, and present the use of Segment Anything (SAM) for this without any finetuning,  i.e., directly utilizing the pre-trained model.
We propose BoxCell, a cell segmentation framework that utilizes SAM's capability to interpret bounding boxes as prompts, \emph{both} at train and test times. At train time, gold bounding boxes given to SAM produce (pseudo-)masks, which are used to train a standalone segmenter.  At test time, BoxCell generates two segmentation masks: (1) generated by this standalone segmenter, and (2) a trained object detector outputs bounding boxes, which are given as prompts to SAM to produce another mask. 
Recognizing complementary strengths, we reconcile the two segmentation masks using a novel integer programming formulation with intensity and spatial constraints. We experiment on three publicly available cell segmentation datasets namely, CoNSep, MoNuSeg, and TNBC, and find that BoxCell significantly outperforms existing box supervised image segmentation models, obtaining 6-10 point Dice gains.
\end{abstract}

\begin{document}

\flushbottom
\maketitle

%
%
\thispagestyle{empty}

\section*{Introduction}
\label{sec:introduction}
Cell segmentation serves as a crucial component for numerous applications, including survival prediction \cite{lu2018nuclear}, tumor/non-tumor classification \cite{wahlby2004combining}, as well as cell counting \cite{tyagi2023degpr}. Our focus is on cell segmentation for histopathology images, which are obtained from tissue biopsies. Generally, cell segmentation models require pixel-level annotations, which are labor-intensive and expensive to obtain.  This is because, typically, many cells are present within a single image, and annotating them requires trained pathologists. Weakly Supervised Image Segmentation (WSIS) addresses this challenge by using weak annotations, which may be present as image-level annotations \cite{ahn2019weakly}, scribbles \cite{lin2016scribblesup}, point annotations \cite{bearman2016s}, or bounding boxes \cite{cheng2023boxteacher, wang2021bounding, tian2021boxinst, mahani2022bounding, yang2023boxsnake}. We study bounding box supervision, as it offers a more accurate estimate of a cell boundary \cite{liu2022box2seg}.

In this work, we explore WSIS with bounding box supervision for cell segmentation in histopathology images, utilizing Segment Anything Model (SAM) \cite{kirillov2023segment}. While SAM has demonstrated remarkable zero-shot performance in various segmentation tasks on natural images, its application to weak supervision, especially for cell segmentation, remains unexplored. We present BoxCell, a SAM-based method for generating segmentation masks using bounding boxes as prompts. 
BoxCell uses SAM in two ways -- at train and test times. At train time, SAM, when prompted with gold bounding box annotations, generates pseudo-masks for training images. These (image, pseudo-mask) pairs supervise the training of a standalone image segmentation model, such as CaraNet \cite{lou2022caranet}. At test time, BoxCell generates two segmentation masks. (1) First mask is generated by this standalone segmenter model. (2) An object detector like YOLO \cite{Jocher_YOLO_by_Ultralytics_2023}, trained with box supervision, predicts bounding boxes on a test image, which are given as prompt to SAM to generate the second mask. Recognizing complementary strengths, with one excelling in localization and the other in learning cell shapes, BoxCell reconciles these strengths using a novel integer programming formulation, with intensity and spatial constraints. Our experiments on three cell segmentation datasets (CoNSep \cite{graham2019hover}, MoNuSeg \cite{verma2020multi}, and TNBC \cite{naylor2018segmentation}) demonstrate BoxCell's significant gains over state-of-the-art weakly supervised segmentation systems, achieving 6-10 point better Dice scores. 
\section*{Related Work}
\label{sec:related_work}
\subsection*{Segment Anything in image segmentation}

Several studies have examined the zero-shot performance of SAM in both natural and medical image segmentation, showing strong results in common scenes but limited performance in complex or low-contrast settings, and on small or irregular objects \cite{ji2024segment,he2023weakly}. Its robustness to image corruptions has also been explored, making it applicable in real-world scenarios \cite{qiao2023robustness}. In medical imaging, SAM has been applied to tasks like liver tumor and brain MRI segmentation \cite{hu2023sam,mohapatra2023sam}. However, in whole slide images (WSIs), SAM performs well on large structures but struggles with small, densely packed cells \cite{deng2025segment}. Recent efforts like Segment Any Cell \cite{na2024segment}, CellSAM \cite{israel2024foundation}, MedSAM \cite{mazurowski2023segment} and $\mu$-SAM\cite{archit2023segment} improve performance by fine-tuning SAM or guiding it with object detectors and box prompts. A common observation across these works is that SAM benefits significantly from prompt-based supervision, particularly with bounding boxes, in challenging medical segmentation tasks.

\subsection*{Weakly supervised image segmentation (WSIS)}
Many studies address weak supervision by scribble \cite{chen2025addressing}, class / attention guided \cite{chen2020deep, chen2024dynamic, chen2024dynamic, chen2022semi} or bounding box supervision for natural image segmentation. Early methods rely on the multi-instance learning (MIL). They assume that bounding boxes are tight \cite{wang2021bounding,kervadec2020bounding}, so a line connecting two opposite edges must contain at least one positive pixel. More recent approaches like BoxInst \cite{tian2021boxinst}, BoxTeacher \cite{cheng2023boxteacher} use box based mask alignment, and BoxSnake \cite{yang2023boxsnake} uses polygon based instance segmentation. Despite the strides made in WSIS for natural images, its progress in cell segmentation remains relatively limited \cite{liu2022weakly}. This limitation can be attributed to challenges like ambiguous boundaries and low contrast variations between foreground and background \cite{chen2022c}.

\subsection*{Ensembling segmentation masks} 
To ensemble multiple segmentation masks, a classic approach involves outputting the average foreground probability for each pixel \cite{ma2021ensembling}. Another method entails creating an ensemble with low precision and high recall, defined by the model diversity metric \cite{ma2021ensembling}. Alternatively, EmergeNet \cite{dai2023samaug} introduces a weighted average of all masks, with weights determined by their performance on the validation set. En-Seg \cite{alush2012ensemble}, use multiple masks to produce an average segmentation masks. We conduct a comparative analysis of BoxCell against these methods (wherever possible), demonstrating BoxCell's superior performance.

\subsection*{Segmentation mask refinement}
Various studies propose post-processing methods to enhance segmentation masks, including GrabCut \cite{rother2004grabcut} and conditional random fields (CRF) \cite{triggs2007scene, plath2009multi}. A popular extension of these works is DenseCRF \cite{krahenbuhl2011efficient}, which uses a fully connected CRF that considers all pairs of pixels in an image. While this approach may be effective on some datasets, DenseCRF typically assumes that all images exhibit consistent and strong contrast between the foreground and background regions, which is often not the case in histopathology images. We conduct the comparative analysis of the BoxCell with DenseCRF and demonstrates BoxCell's superior performance.

\begin{figure*}[t]
    \centering
    \includegraphics[width = \linewidth]{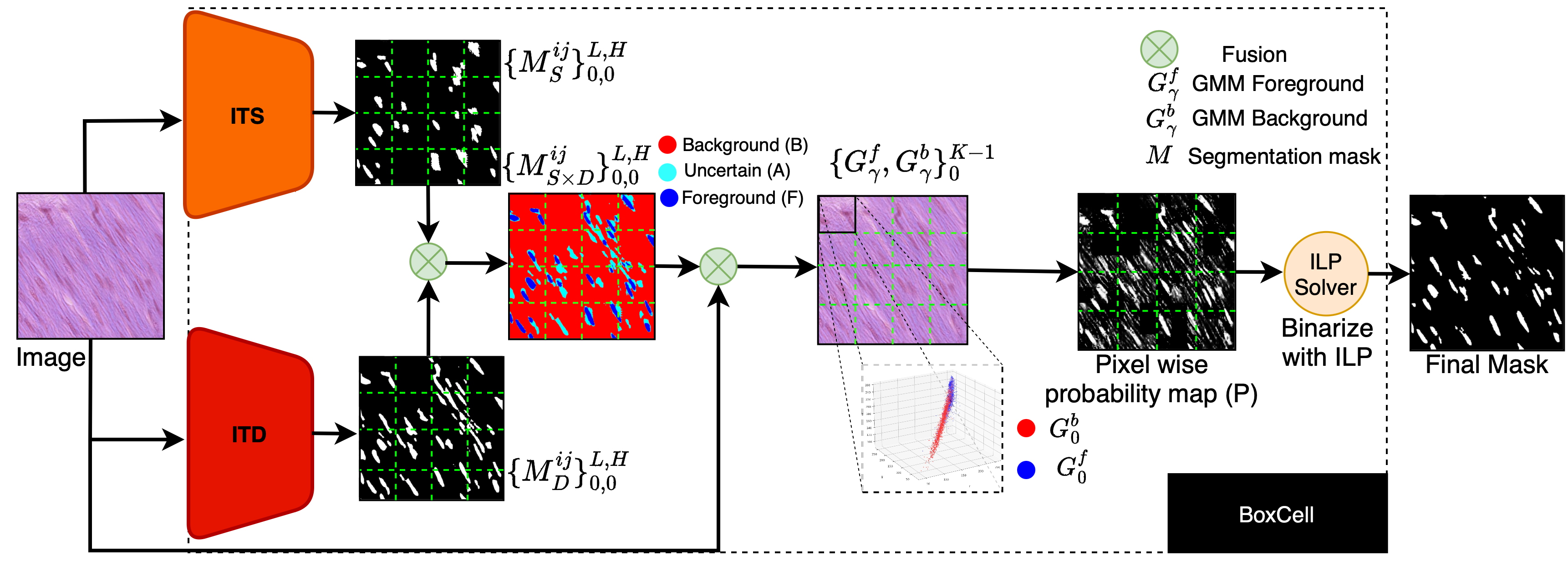}
    \caption{Inference pipeline for BoxCell, which produces masks  $M_{D}$ and $M_{S}$ using a ITD and ITS. These masks are split into a $K$$\times$$K$ grid, and GMMs are trained to estimate probability map ($P$). ILP solver refines $P$ based on intensity and spatial constraints. This figure was created using draw.io.}
    \label{fig:Model Diagram}
\end{figure*}
\section*{BoxCell: Our Proposed Method}
\label{sec:method}

Weakly supervised image segmentation (WSIS) takes in training dataset \(D = \{X^T_{k}, B^T_{k}\}_{k=1}^{D}\), where \(X^T_{k}\) is a training image, and \(B^T_{k}\) (in our setting) represents bounding box annotations for the target class. Its goal is to train a model, which, given a test image \(X\), predicts a foreground (cell) segmentation mask \(M\). Our proposed method, BoxCell, has two components, Inference Time Segmenter (ITS) and Inference Time Detector (ITD), which both employ SAM -- a prompting based general-purpose segmentation model -- at train and test times, respectively (see Fig \ref{fig:Model Diagram}). The proposed method, \textit{BoxCell}, consists of two components: the \textit{Inference Time Detector (ITD)} and \textit{Inference Time Segmenter (ITS)}. Both components leverage SAM, a prompt-based general-purpose segmentation model, operating at both training and test stages (see Fig.~\ref{fig:Model Diagram}).

ITD and ITS operate on test image \(X\) and independently generate segmentation masks, \(M_D\) and \(M_S\). We find that the two masks possess complementary strengths, and better results can be achieved by merging the two. BoxCell achieves this via a novel Integer Linear Programming (ILP) formulation. The ILP outputs a final mask \(M\) by balancing the probability of pixel classification into classes (foreground and background), along with the goal that similar intensities at neighboring pixels should be assigned the same class.

\subsection*{Generating Segmentation Masks}

\begin{figure*}[h!]
    \centering
    \includegraphics[scale = 0.23, width = 17.5cm, height = 5cm]{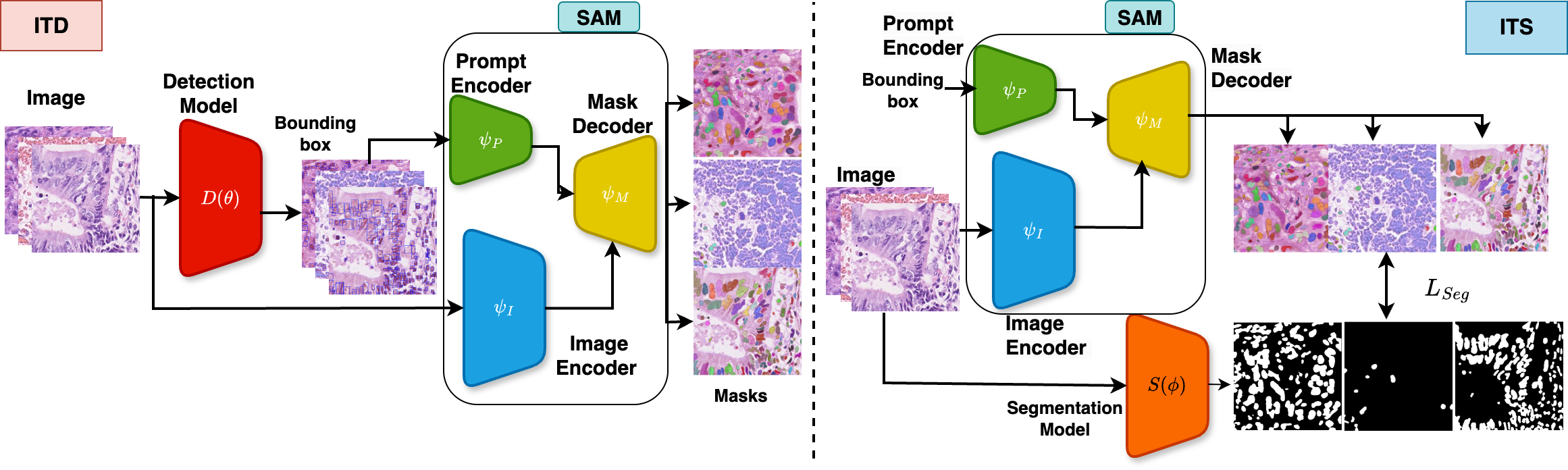}
    \caption{Workflows with SAM in weak supervision. ITD uses the detection model $D(\theta)$ to predict bounding boxes. The detection model is trained on the training data and is used to predict bounding boxes, which are used as box prompts for SAM during inference. ITS uses segmentation masks predicted by SAM as pseudo ground truth to train $S(\phi)$. We only call $S(\phi)$ during inference.}
    \label{fig:Model Diagram_ITD_ITS}
\end{figure*} 

\subsubsection*{Inference time detector (ITD): }

ITD (see Fig. \ref{fig:Model Diagram_ITD_ITS}), trains an object detector \(D(\theta)\) such as Yolov8 \cite{Jocher_YOLO_by_Ultralytics_2023} using images \(X^T_{k}\) and the set of gold bounding boxes \(B^T_{k}\).
The object detector is trained with objectness, classification, and localization losses. Objectness loss (\(L_{obj}\)) is the confidence score indicating whether the box contains an object or not. Classification loss (\(L_{cls}\)) is computed as the binary cross-entropy between the predicted class and ground truth class. Localization loss (\(L_{loc}\)) is the error in predicted bounding box coordinates as compared to ground truth bounding box coordinates. The total detection loss is the sum of all three losses, given as \(L_{det} = L_{obj} + L_{cls} + L_{loc}\). This object detector \(D(\theta)\) predicts a set of bounding boxes \(\hat{B}\) for a test image \(X\). Each box \(b \in \hat{B}\) is used as a prompt to SAM to generate a segmentation mask within that box. All box-level masks are combined to generate the image-level segmentation mask \(M_D\).


\subsubsection*{Inference Time Segmentor (ITS):}

During training, SAM generates masks \(M^T_{k}\) for the training images (\(X^T_{k}\)) using ground truth bounding boxes \(B^T_{k}\) as prompts. Despite SAM's errors, these masks serve as pseudo-masks for training a standalone segmentation model \(Sg(\phi)\) like CaraNet \cite{lou2022caranet} -- it is trained using a sum of Dice loss and BCE: \(L_{\text{seg}} = L_{\text{bce}} + L_{\text{dice}}\) (see Fig. \ref{fig:Model Diagram_ITD_ITS}). Binary cross-entropy (\(L_{bce}\)) improves the pixel-level classification of the segmentation mask, and Dice loss (\(L_{dice}\)) guides the intersection of the prediction with the ground truth masks, thereby improving the localization of the predicted masks. At test time, \(Sg(\phi)\) runs on \(X\) to directly generate an image-level segmentation mask, \(M_S\).

\subsection*{Integer Programming for Reconciling Segmentation Masks}


We find that $M_D$ excels in localization, whereas $M_S$ is better at shapes; BoxCell merges the two for better performance. We make two key observations. First, for histopathological images, the intensity values within pixels of one class (foreground or background) vary significantly, due to variations in tissue structure and amount of staining from one part of image to another. So, any intensity distribution learned over the \emph{whole} image is likely to be noisy, but could be meaningful if learned over a small patch of the image. Second, there still exists perceptible contrast between pixel intensities in the vicinity of the boundary of a segmentation mask. Following these observations, BoxCell learns Gaussian Mixture Models (GMMs) to model \emph{patch-level} intensity distributions for foreground and background. It then casts an ILP that maximizes the GMM prediction probability along with a soft constraint that neighboring pixels are assigned different classes only if their intensity difference is high. 
To do so, BoxCell divides each pixel $(i,j)$ for a test image $X$ into three sets: $F$, $B$ and $A$. Here, $F$ (and $B$) is the set where masks $M_D$ and $M_S$ agree on the pixel to be in foreground (resp., background); and $A$ is where they disagree, i.e., $A=\{(i,j)~|~M_D(i,j)\neq M_S(i,j)\}$. BoxCell accepts the pixel labels for $F$ and $B$ for final mask $M$ and only attempts to reassign labels in $A$.

\subsection*{Learning GMMs}

BoxCell splits the image $X$ of size $L\times H$ into $K^2$ mutually exclusive and collectively exhaustive patches of size $L/K \times H/K$ each. For each patch $\gamma$, it learns two GMMs, one for foreground pixel intensities, and one for background. It uses pixels in $F \cup B$ to learn these GMMs and ignores ambiguous pixels in the patch. 
More formally, $G^{f}_{\gamma}$ and $G^{b}_{\gamma}$ are $N$-component 3-dimensional (RGB) GMMs over the foreground and background pixels in a patch $\gamma$. Let $w = \{w_1,w_2...w_N\}$ be the mixture weights such that $\sum w_n=1$ and $0\leq w_n \leq 1$. Let $\mu = \{\mu_1,\mu_2...\mu_N\}, \mu_n \in \mathbb{R}^3$  
and $\Sigma = \{\Sigma_1,\Sigma_2...\Sigma_N\}$, $\Sigma_n = [\sigma^2]_{3\text{x}3}$, respectively, denote the means and co-variances. The likelihood density of an RGB pixel $c = (c^1, c^2, c^3)$ belonging to a mixture $G$ is given by $p'(c ~| \ G; \mu, \Sigma) = \sum_{n=1}^{N}w_n N(c, \mu_n,\Sigma_n)$, where $N$ is the Gaussian function.

\begin{equation}
 \frac{1}{(2\pi)^{1.5}|\Sigma_n|^{0.5}}\exp \left (-\frac{(c-\mu_n)^T\Sigma_n^{-1}(c-\mu_n)}{2}\right )
\label{eq:multivariate_gauss}
\end{equation} 
Since each pixel can either belong to foreground or background, we normalise probabilities as

\begin{equation}
p(c ~| \ G^{f}_{\gamma}) = \frac{p'(c ~| \ G^{f}_{\gamma})}{p'(c ~| \ G^{f}_{\gamma}) + p'(c ~| \ G^{b}_{\gamma})}
\label{eq:normalised_prob}
\end{equation} 

\noindent Here, $p(c~|~G^{f}_{\gamma})$ is the probability of pixel $c$ being in the foreground, and $p(c~|~G^{b}_{\gamma}) = 1 - p(c~|~G^{f}_{\gamma})$ of it being in the background. Note that these probabilities are solely based on pixel intensities and do not incorporate any spatial information. BoxCell merges $p(c~|~G^{f}_{\gamma})$ for all the patches $\gamma$ to create a complete probability distribution over the entire image -- we denote it as $P(c)$ for RGB pixel $c$.

\subsection*{Integer Linear Programming}
For a pixel $(i,j)$, lets its color information (RGB) be denoted as $c_{ij}$ (a 3-tuple).
The ILP first defines a binary variable $x_{ij}$ (for ambiguous pixels), which is $1$, iff the pixel is assigned the foreground label. It defines a part of the objective function, $O_{idf}$, where $idf$ stands for Intensity Distribution Factor:
\begin{equation}
O_{idf} = {\sum_{(i,j)\in A}x_{ij}P(c_{ij}) + (1-x_{ij})(1-P(c_{ij}))}
\label{eq:idf}
\end{equation}

To ensure well-formedness of cells, ILP imposes that neighboring pixels that are assigned different labels must differ in their intensities. For this, it defines binary edge variables $e_{ij0}$ and $e_{ij1}$, which encode the edges between pixels $(i,j)$ and $(i+1,j)$, and between $(i,j)$ and $(i,j+1)$, respectively. 
The edge variables are assigned 0 if both pixels on the edge belong to the same class, and 1 otherwise. This is encoded in constraints as $e_{ij0} = |x_{ij}-x_{(i+1)j}|$ and $e_{ij1} = |x_{ij}-x_{i(j+1)}|$. If two neighboring pixels get different labels, the objective function gets penalized based on their intensity differences: 
\begin{equation}
O_{scf} = \sum_{i=1}^{L-1}\sum_{j=1}^{H}e_{ij0}S_{ij0} + \sum_{i=1}^{L}\sum_{j=1}^{H-1}e_{ij1}S_{ij1}
\label{eq:scf}
\end{equation}
We name this part of objective as $O_{scf}$, where $scf$ stands for Spatially Constraining Factor. Here, $S_{ij0}$ and $S_{ij1}$ are a function of intensity differences, for which we employ the color similarity metric \cite{tian2021boxinst}, with $\theta$ as a hyperparameter:
\begin{equation}
S_{ij0} = exp\Bigg(\frac{-||c_{ij}-c_{(i+1)(j)}||_2}{\theta}\Bigg),
\label{eq:color0}
\end{equation} 
\begin{equation}
S_{ij1} = exp\Bigg(\frac{-||c_{ij}-c_{(i)(j+1)}||_2}{\theta}\Bigg).
\label{eq:color1}
\end{equation}
Overall, the complete ILP formulation is as follows, with $x_{ij}$ values computing the final segmentation mask labels in $M$ for pixels in set $A$: 

\begin{equation}
\begin{gathered}
\underset{x_{ij}, e_{ij0}, e_{ij1}}{\text{maximize}}
 O_{idf} - \lambda O_{scf} \\
\text{subject to} \\
 e_{ij0} = |x_{ij} - x_{(i+1)j}|, \\
 e_{ij1} = |x_{ij} - x_{i(j+1)}|, \\
 x_{ij}, e_{ij0}, e_{ij1} \in \{0,1\}.
\end{gathered}
\end{equation}

\begin{figure*}[t]
    \centering
    \includegraphics[scale = 0.5, , width = 16cm, height = 7cm]{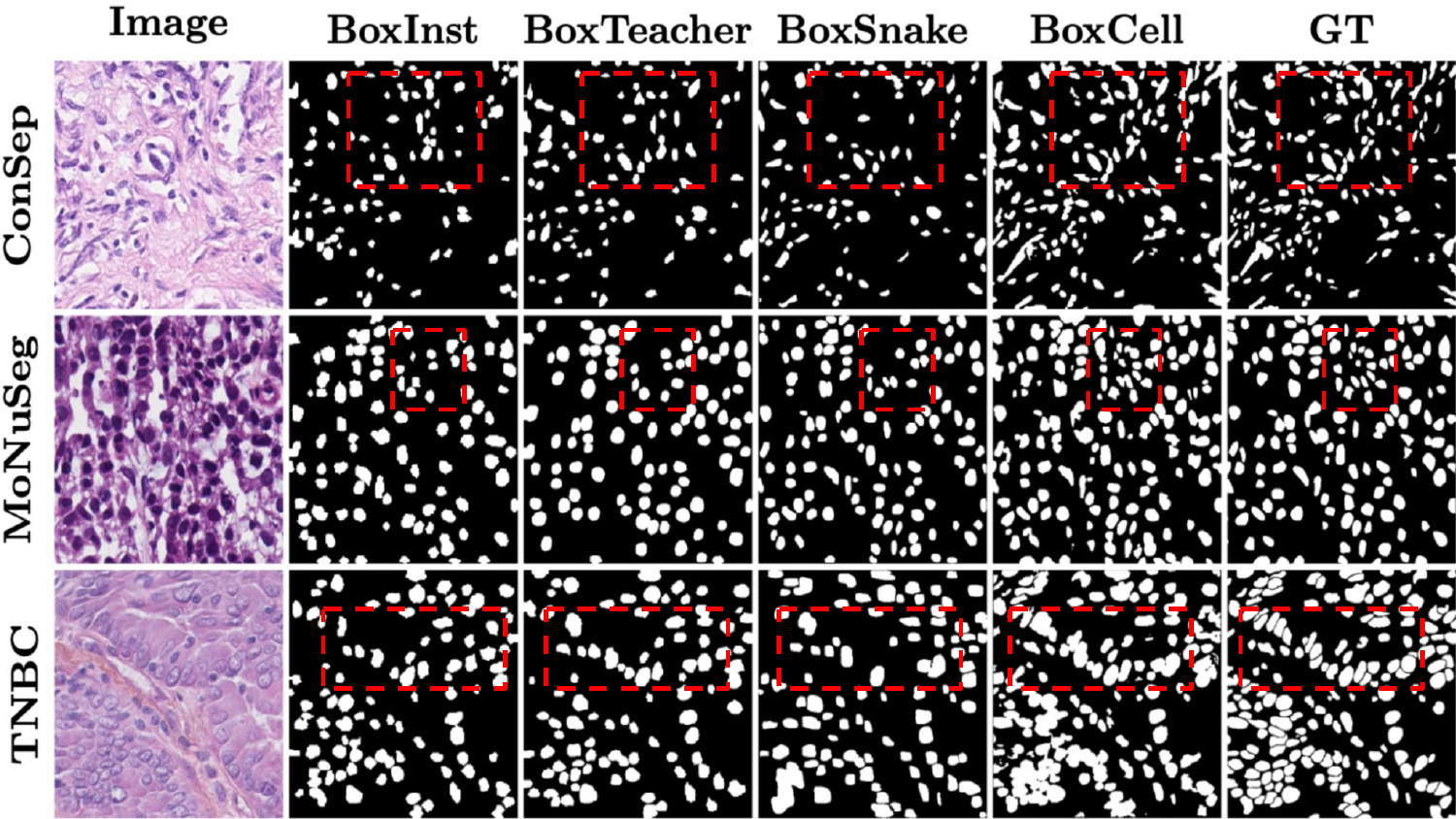}
    \caption{Qualitative analysis of segmentation masks. Column 1 is the original image, Columns 2-5 show cropped masks (shown in red box) generated from three comparison models and BoxCell. Last column is the ground truth. BoxCell exhibits best results, providing more accurate masks with better cell boundary and shape.}
    \label{fig: Auxillary-SAM}
\end{figure*}

\section*{Results}
\label{sec:results}

The primary goal of our experiments is to compare BoxCell's performance with existing box-supervised segmentation methods. Moreover, we wish to understand the qualitative differences between ITD and ITS masks, if any. Finally, we also compare BoxCell's ILP formulation against existing mask merging and 
mask refinement approaches
. 

\subsection*{Datasets}
\textbf{CoNSep}: Colorectal nuclear segmentation and phenotypes (CoNSeP) \cite{graham2019hover} is a nuclear segmentation and classification dataset of H\&E stained images. Each image is of 1000$\times$1000 dimension. The dataset deals with single cancer, and colorectal adenocarcinoma (CRA) images. It consists of a total of 41 whole slide images (WSI), which have a total of 24,319 annotated cells of 3 classes: inflammatory cells, epithelial cells, and spindle cells. A total of 27 images are used for training and validation, and the rest 14 are used for testing. Further, we split the 1000×1000 images into four sub-images of dimension 500×500. This results in a dataset of 98 train, 10 validation and 56 test images.\\
\textbf{MoNuSeg}: Multi-organ nuclei segmentation (MoNuSeg) \cite{kumar2019multi} is a nuclei segmentation dataset of H\&E images representing cell nuclei from 7 different organs like breast, liver, kidney, prostate, bladder, colon and stomach to ensure diversity of nuclear appearances. It consists of a total of 51 images containing 28846 annotated cells. A total of 37 are used for training and validation, and 14 are used as test images. We split the 1000x1000 image into four 500x500 images, resulting in 133 train, 15 validation and 56 test images.\\
\textbf{TNBC}: It consists of H\&E slides of triple negative breast cancer patients taken at 40x magnification \cite{naylor2018segmentation}. The dataset contains 50 images with a total of 4022 annotated cells. Each image is of 512x512 dimension. This dataset proves valuable for evaluating model performance under varying degrees of cellularity. Out of 50 images, we use 34 as training samples, 5 for validation and 11 for the test set.\\
Since all datasets are originally image segmentation datasets, we converted instance segmentation masks into bounding boxes and used them to train our object detector, keeping the golden masks held out at the time of training. Some of these datasets also assign classes to the cells, but for the sake of our problem, we are only interested in binary segmentation and ignore the class labels.

\subsection*{Evaluation metrics}
To evaluate BoxCell's performance on the semantic segmentation task, we use the Dice coefficient and Intersection over Union (IoU) metrics to quantify the similarity between the predicted and ground truth masks.

\subsection*{Instance segmentation}
\label{sec:Instance segmentation}
Although BoxCell generates semantic segmentation masks, we convert these into instance segmentation masks to evaluate performance alongside instance segmentation methods. This conversion uses the outputs from both ITD and BoxCell segmentation masks. While generating ITD mask, each cell is assigned a unique instance ID. We then check for overlaps between the ITD and BoxCell masks, and in regions where overlap exists, we assign the corresponding ITD instance ID to those pixels of BoxCell mask. For regions predicted by BoxCell but not by ITD, we assign unique instance IDs to ensure they are included in the instance segmentation mask evaluation. In cases where the BoxCell mask contains connected cells, we apply k-means clustering to separate them into distinct instances. Additionally, we evaluate the instance segmentation quality using three metrics: Aggregated Jaccard Index (AJI) \cite{kumar2019multi}, Panoptic Quality (PQ) \cite{graham2019hover}, and Boundary F1-score (BF1) \cite{csurka2013good}. Panoptic Quality (PQ) combines detection quality (DQ) and segmentation quality (SQ), offering a comprehensive measure of both the accuracy of object detection and the precision of segmentation. In our baselines, WSIS methods, except for SPN, directly produce instance segmentation masks. For ensembling and mask refinement methods, we follow the same process used in BoxCell to generate instance segmentation masks.

\subsection*{Implementation details}

For the object detection component (ITD), we employ YOLOv8x~\cite{Jocher_YOLO_by_Ultralytics_2023} trained for 300 epochs with early stopping based on validation performance. The initial learning rate is set to 0.01 and gradually reduced by a factor of 0.01 during training using a cosine decay schedule with a 5-epoch warmup. We use a batch size of 32. Data augmentations include random horizontal and vertical flips, rotations ($\pm90^\circ$), and color jitter (brightness/contrast $\pm0.2$). During inference, we apply a detector confidence threshold of 0.3 and an NMS IoU threshold of 0.5. All experiments are conducted on NVIDIA RTX-5000 and Tesla A100 GPUs.

For the auxiliary segmentation stage (ITS), we adopt CaraNet~\cite{lou2022caranet} and SegFormer \cite{xie2021segformer}, which utilizes a reverse axial attention mechanism effective for small or tiny object segmentation. CaraNet is trained for 200 epochs using the AdamW optimizer with an initial learning rate of $1\times10^{-4}$, cosine decay scheduling, and a 5-epoch warmup. We set the batch size to 16 and apply early stopping with a patience of 20 epochs based on validation Dice. The same data augmentations as ITD are used, and during inference, small objects with an area smaller than 20 pixels are removed. For comparison, we also evaluate BBTP++~\cite{wang2021bounding}, BoxInst~\cite{tian2021boxinst}, BoxSnake~\cite{yang2023boxsnake}, BoxTeacher~\cite{cheng2023boxteacher}, and SPN~\cite{liu2022weakly}, all trained under identical settings for fairness.

To reconcile the predictions from ITD and ITS, we propose an Integer Linear Programming (ILP)-based mask fusion strategy. We compare our ILP method against several existing mask merging strategies. (i) \textit{AP (Averaging)} performs element-wise averaging of ITD and ITS masks, normalized by 2 and binarized at a threshold of 0.5. (ii) \textit{LP (Low Precision Averaging)}~\cite{ma2021ensembling} applies the same averaging but uses a higher binarization threshold of 0.9 to emphasize high-confidence regions. (iii) \textit{ENet}~\cite{das2023emergenet} computes a weighted sum of ITD and ITS masks, where the weights are tuned on the validation set and remain balanced across datasets (e.g., CoNSeP: ITD=0.522 vs. ITS=0.477; MoNuSeg: ITD=0.523 vs. ITS=0.476; TNBC: ITD=0.493 vs. ITS=0.507). (iv) \textit{ILP (ours)} formulates the mask reconciliation as an integer linear programming problem, achieving consistent improvements across datasets, as shown in Table~\ref{tab:Mask_Merging_with_time}.

To reconcile the predictions from ITD and ITS, we employ an Integer Linear Programming (ILP)-based mask fusion strategy. We compare ILP with several alternatives: AP (Averaging) performs element-wise averaging of ITD and ITS masks followed by binarization at 0.5; LP (Low Precision Averaging) ~\cite{ma2021ensembling} uses a higher threshold of 0.9 to emphasize confident regions; and \textit{ENet}~\cite{das2023emergenet} computes a weighted sum of ITD and ITS masks with validation-tuned weights that remain balanced across datasets.

Following mask fusion, we refine the results using DenseCRF post-processing with Pairwise Gaussian (size $3\times3$) and Pairwise Bilateral (size $5\times5$) kernels for 10 inference iterations. For BoxCell-ILP solver, each image is partitioned into a $K\times K$ grid for RGB-GMM modeling, where we use $K=5$ in all experiments. Hyperparameters were tuned via grid search using Gurobi~\cite{gurobi}, with $\lambda \in \{0.5,1,2,5\}$ (pairwise smoothness weight), $\theta \in \{1,5,10,25,30\}$ (color similarity scaling), and the number of GMM components $\in \{2,3,4,5,6\}$. The best results were obtained with $\lambda=2$, $\theta=25$, and 2 GMM components. For an image of size $500\times500$, the complete processing time is approximately 67--71 seconds on an Intel Xeon processor with 32 CPU cores.


\begin{table*}[t!]
    \centering
    
    \begin{tabular}{p{180pt} p{33pt} p{33pt} p{33pt} p{33pt}  p{33pt} p{33pt}}
    \toprule
        Model  & \multicolumn{2}{c}{CoNSep} & \multicolumn{2}{c}{MoNuSeg} & \multicolumn{2}{c}{TNBC} \\ \hline
        & Dice & IoU & Dice & IoU & Dice & IoU \\
    \midrule
        SPN \cite{liu2022weakly}  & 64.44 & 60.54 & 50.48 & 35.29 & 68.88 & 53.68 \\
        BoxSnake \cite{yang2023boxsnake}  & 71.02 & 55.40 & 74.80 & 60.58 & 69.71 & 56.32 \\
        BoxTeacher \cite{cheng2023boxteacher}  & 63.42 & 43.28 & 69.64 & 54.81 & 74.35 & 59.32 \\
        BBTP \cite{hsu2019weakly}  & 70.89 & 55.74 & 72.46 & 60.64  & 68.53 & 58.14 \\ 
        BBTP++ \cite{wang2021bounding}  & 75.58 & 60.99 & 72.84 & 61.22  & 70.37 & 58.60 \\
        BoxInst \cite{tian2021boxinst} & 64.50 & 48.11  &66.7 & 52.16  & 74.75 & 59.89\\
        SAM-BBTP \cite{hsu2019weakly} & 68.62 & 52.78 & 69.99 & 53.99 & 78.78 & 65.04 \\
        SAM-BBTP$++$ \cite{wang2021bounding} & 74.64 & 60.11 & 73.58 & 60.64 & 79.64 & 67.28 \\
        SAM-BoxInst \cite{tian2021boxinst} & 74.08 & 59.43 & 73.05 & 61.22 & 79.43 & 67.43 \\
        ITD (Ours) & 80.00 & 66.99 & 79.87 & 66.68 & 82.86 & 70.71 \\
        ITS (Ours) & 79.86 & 65.54 & 79.38 & 65.97 & 80.66 & 66.18 \\
        BoxCell - DenseCRF (Ours) \cite{krahenbuhl2011efficient} & 77.82 & 63.50 & 80.41 & 67.40 & 81.19 & 68.50 \\
        BoxCell - ILP (Ours) & \textbf{81.39} & \textbf{68.82} & \textbf{81.74} & \textbf{69.25} & \textbf{85.01} & \textbf{74.06} \\
    \bottomrule
    \end{tabular}
    \caption{Comparison of Bounding Box Supervised Methods}
    \label{tab:Auxiliary_model}
\end{table*}

\subsection*{Comparison of Weak Supervised Approaches}
\subsubsection*{Semantic Segmentation}


Table \ref{tab:Auxiliary_model} compares all models in the weakly supervised image segmentation setting. BoxCell achieves substantial 6-10 point Dice improvements compared to the strongest non-SAM competitors across datasets, demonstrating the significant merit of our approach. Even when competing against SAM-based methods, BoxCell maintains a consistent 6-7 point Dice advantage. We believe that this is because the heuristics imposed by weak supervision losses are insufficient to guide SAM. For instance, BoxInst employs intensity-dependent losses in local neighborhoods that remain static across image patches, ignoring underlying image distribution variations. Similarly, SAM-BBTP++ lacks intensity-based criteria for mask integrity and fails to penalize contradictory predictions in neighborhoods. In preliminary experiments, we also made other attempts to use weak supervision losses for training SAM and found that they generally confuse SAM (because of catastrophic forgetting). Figure \ref{fig: Auxillary-SAM} illustrates sample predictions from each dataset, where BoxCell demonstrates superior accuracy in capturing cell boundaries and shapes.

We also compare with mask refinement method like DenseCRF. DenseCRF uses a unary classifier to learn long-term dependencies. However, we observed that in cases where the background varies throughout the image and the contrast between the foreground and background is minimal, such as in histopathology images, DenseCRF produces suboptimal performance. Therefore, learning local features proves to be more beneficial as done in BoxCell. BoxCell achieves, 1 $-$ 4 pt dice gain as compared to DenseCRF. Closest to our work, ENSeg-ILP focuses on averaging segmentation results, whereas BoxCell aims to maximize accuracy by reconciling two complementary predictions—ITD and ITS—resulting in notable Dice score improvements (ITD: 82~$\rightarrow$~85, ITS: 80~$\rightarrow$~85). In addition, BoxCell introduces spatial constraints based on pixel color, contributing an additional 1.1 - 1.5 Dice points compared to using intensity constraints alone, as employed by ENSeg-ILP (see Table~\ref{tab:Ablation}). Although the code for ENSeg-ILP is not publicly available for a direct comparison, BoxCell's enhancements highlight its clear advantage. 


As described in the dataset details, each ConSeP and MoNuSeg WSI ($1000\times1000$) was divided into four $500\times500$ sub-images for evaluation, with per-WSI analysis showing negligible differences (ConSeP: 81.39 vs.\ 81.41; MoNuSeg: 81.74 vs.\ 81.73). TNBC was evaluated at full resolution. We performed paired t-test analysis to evaluate the statistical significance of BoxCell-ILP against existing bounding box supervised approaches. Statistical analysis confirms that BoxCell achieves significant improvements ($p < 10^{-4}$), with 95\% confidence intervals for mean Dice gains of [5.7–5.8], [8.1–8.2], and [5.2–5.3] on ConSeP, MoNuSeg, and TNBC, respectively, validating its robustness across datasets.


Lastly, we compare BoxCell with several SAM variants (Table \ref{tab:SAM_backbone}), including the recently introduced SAM2~\cite{ravi2024sam2}, designed for video segmentation, and MedSAM~\cite{ma2024segment}, trained specifically on medical images. We also include $\mu$-SAM, a SAM variant optimized for microscopy images. As shown in Table \ref{tab:SAM_backbone}, BoxCell consistently improves ITD-ITS performance across all SAM backbones. For instance, on the CoNSeP dataset, BoxCell improves performance from 79.97/79.80 to 81.36 with SAM2, and on MoNuSeg, from 80.35/79.80 to 81.83. On TNBC, it increases from 82.24/80.54 to 84.46. Similarly, with MedSAM, the Dice score rises from 71.00 to 74.91, though its overall performance remains lower than SAM and SAM2 for cell segmentation. Quantitatively, BoxCell achieves improvements of approximately $+6.5$ Dice on CoNSeP, $+4.5$ Dice on MoNuSeg, and $+2.3$ Dice on TNBC over $\mu$-SAM and MedSAM. These results highlight that beyond simply adapting SAM to biomedical images, BoxCell’s integer linear programming–based reconciliation of inference-time detection and segmentation yields consistent 2--3 Dice point gains across backbones and substantial absolute improvements overall.

\begin{table*}[t]

\centering
\begin{tabular}{p{80pt} p{20pt} p{20pt} p{20pt} p{20pt} p{20pt} p{20pt}}
\toprule
Method & \multicolumn{2}{c}{CoNSep} & \multicolumn{2}{c}{MoNuSeg} & \multicolumn{2}{c}{TNBC} \\ \hline 
BoxCell     & Dice & IoU & Dice & IoU & Dice & IoU \\ 
\midrule
with SAM \cite{kirillov2023segment} & \textbf{81.39} & \textbf{68.82} & 81.74 & 69.25 & \textbf{85.01} & \textbf{74.06} \\
with SAM2 \cite{ravi2024sam2} & 81.36 & 68.80 & \textbf{81.83} & \textbf{69.35} & 84.46 & 73.61 \\
with MedSAM \cite{ma2023segment} & 74.91 & 60.83 & 77.60 & 64.79 & 82.76 & 70.72 \\
with $\mu$-SAM \cite{archit2025segment} & 74.94 & 60.92 &77.22 & 64.35 &81.07 & 68.16 \\
\bottomrule
\end{tabular}
\caption{Comparison of BoxCell with various SAM backbones}
\label{tab:SAM_backbone}
\end{table*}

\subsubsection*{Instance Segmentation}

Table~\ref{tab:instance-det_dependence} compares BoxCell with detector-agnostic instance conversion methods, including connected components and watershed segmentation. In our current pipeline, the number of clusters ($K$) for splitting touching regions is determined by the number of overlapping bounding boxes from the detector, with centroids initialized at corresponding box centers. This introduces a degree of dependence on the detector. To assess its impact, we implemented detector-agnostic alternatives (connected-component and watershed) and found that, while they achieve reasonable performance, BoxCell’s detector-based instance conversion consistently yields substantially higher PQ and AJI scores across all datasets. 

\begin{table}[h!]
\centering
\begin{tabular}{l|cc|cc|cc}
\hline
\multirow{2}{*}{Method} & \multicolumn{2}{c|}{ConSep} & \multicolumn{2}{c|}{MoNuSeg} & \multicolumn{2}{c}{TNBC} \\
 & PQ & AJI & PQ & AJI & PQ & AJI \\
\hline
Connected Components & 28.75 & 22.65 & 53.90 & 52.55 & 50.70 & 40.78 \\
Watershed            & 30.16 & 27.55 & 57.50 & 58.13 & 53.70 & 49.60 \\
BoxCell (ours)       & \textbf{42.75} & \textbf{44.72} & \textbf{62.51} & \textbf{63.75} & \textbf{63.90} & \textbf{66.26} \\
\hline
\end{tabular}
\caption{Comparison of detector-agnostic instance conversion methods with BoxCell.}
\label{tab:instance-det_dependence}
\end{table}

Furthermore, Table~\ref{tab:Instance Segmentation} compares BoxCell with other baselines on the instance segmentation task. Although BoxCell is primarily designed for semantic segmentation, we extend it to instance segmentation and observe consistent improvements in PQ, AJI, and Boundary F1. BoxCell achieves 2--7 point PQ gains over weakly supervised instance segmentation methods, particularly excelling in densely packed or low-contrast regions where competing approaches often struggle.

\begin{table*}[t]
\centering
\begin{tabular}{p{120pt} p{28pt} p{28pt} p{28pt} p{28pt}  p{28pt} p{28pt} p{28pt} p{28pt} p{28pt}}
\hline
\textbf{Model}       & \multicolumn{3}{c}{\textbf{CoNSep}} & \multicolumn{3}{c}{\textbf{MoNuSeg}} & \multicolumn{3}{c}{\textbf{TNBC}} \\ \hline
                     & \textbf{PQ} & \textbf{AJI} & \textbf{BF1} & \textbf{PQ} & \textbf{AJI} & \textbf{BF1} & \textbf{PQ} & \textbf{AJI} & \textbf{BF1} \\ \hline

BoxTeacher \cite{cheng2023boxteacher} & 34.66       & 23.95        & 51.09       & 42.65       & 42.05       & 49.92       & 47.24       & 46.67       & 52.33       \\
BoxSnake \cite{yang2023boxsnake} & 33.7        & 34.87        & 51.60       & 56.43       & 56.51       & 51.18       & 53.71       & 52.64       & 53.02       \\ 
BoxInst   \cite{tian2021boxinst}           & 41.63       & 28.5         & 52.43       & 55.01       & 55.5        & 53.90       & 56.01       & 56.79       & 51.78       \\ 
SAM-BBTP \cite{hsu2019weakly} & 35.51 & 27.39 & 51.12& 48.32 & 48.83 & 50.6 & 51.48 & 52.09 & 52.5\\
SAM-BBTP++ \cite{wang2021bounding}  & 41.97 & 29.65 & 52.38 & 56.07 & 56.91 & 53.92& 56.13 & 56.87 & 52.56\\
AP (ITS, ITD) \cite{ma2021ensembling} & 39.90 & 28.41 & 50.18 & 54.17 & 54.98 & 52.21 & 55.32 & 55.12 & 51.73\\
LP (ITS, ITD) \cite{ma2021ensembling} & 40.01 & 28.56 & 50.66 & 54.91 & 52.50 & 52.12 & 54.42 & 54.86 & 51.23\\
ENet (ITS, ITD) \cite{das2023emergenet} & 39.84 & 28.35 & 50.02 & 54.12 & 54.94 & 52.17 & 55.29 & 55.08 & 51.67 \\
BoxCell-DenseCRF (Ours) \cite{krahenbuhl2011efficient} & 35.03       & 40.66        & 53.87       & 59.01       & 60.51       & 53.62       & 50.61       & 58.02       & 52.92       \\ 
BoxCell - ILP (Ours)             & \textbf{42.75}       & \textbf{44.72}        & \textbf{54.07}       & \textbf{62.51}       & \textbf{63.75}       & \textbf{54.03}       & \textbf{63.9}        & \textbf{66.26}       & \textbf{53.85}       \\ \hline
\end{tabular}
\caption{Performance Comparison for Instance Segmentation}
\label{tab:Instance Segmentation}
\end{table*}

\subsubsection*{Comparison with mask merging methods}

Table~\ref{tab:Mask_Merging_with_time} reports experiments where masks $M_D$ and $M_S$ from ITD and ITS are merged. We compare several mask merging strategies from the literature and our proposed ILP-based approach. AP (Averaging) performs element-wise averaging of ITD and ITS masks, followed by normalization by 2 and binarization at a threshold of 0.5. LP (Low Precision Averaging) \cite{ma2021ensembling} uses the same averaging but applies a higher threshold of 0.9 to emphasize high-confidence regions. ENet \cite{das2023emergenet} computes a weighted sum of ITD and ITS masks, where the weights are tuned on the validation set. The weights are nearly balanced across datasets, indicating stability (e.g., CoNSeP: ITD=0.522 vs. ITS=0.477; MoNuSeg: ITD=0.523 vs. ITS=0.476; TNBC: ITD=0.493 vs. ITS=0.507). Finally, ILP (ours) reconciles ITD and ITS masks through integer linear programming. Across all datasets, BoxCell’s ILP achieves consistent improvements over them.

        

\begin{table*}[h!]
    \centering
    \begin{tabular}{p{140pt} p{60pt} p{33pt} p{33pt} p{33pt} p{33pt} p{33pt} p{33pt}}
    \toprule
        \textbf{Model} & \textbf{Time (sec)} & \multicolumn{2}{c}{\textbf{ConSep}} & \multicolumn{2}{c}{\textbf{MoNuSeg}} & \multicolumn{2}{c}{\textbf{TNBC}} \\ \hline
        & & Dice & IoU & Dice & IoU & Dice & IoU \\
    \midrule
        AP (ITS, ITD) & 1.20 & 80.55 & 67.69 & 80.04 & 66.86 & 82.92 & 70.97 \\
        LP (ITS, ITD) & 1.26  & 80.59 & 66.70 & 79.57 & 66.37 & 78.48 & 70.07 \\
        ENet (ITS, ITD) & 2.63  & 80.23 & 67.40 & 80.07 & 66.59 & 82.26 & 71.07 \\
        BoxCell - Sparse & 15.00 & 80.15 & 67.10 & 80.21 & 67.02 & \textbf{85.02}& \textbf{74.07} \\
        BoxCell - Gurobi & 69.02 & \textbf{81.39} & \textbf{68.82} & \textbf{81.74} & \textbf{69.25} & 85.01 & 74.06 \\
    \bottomrule
    \end{tabular}
    \caption{Comparison of Mask Merging Methods.}
    \label{tab:Mask_Merging_with_time}
\end{table*}

\begin{table}[t]

\centering
\begin{tabular}{p{60pt} p{15pt} p{15pt} p{15pt} p{15pt} p{15pt} p{15pt}}
\toprule
Method & \multicolumn{2}{c}{ConSep} & \multicolumn{2}{c}{MoNuSeg} & \multicolumn{2}{c}{TNBC} \\ \hline 
       & Dice & IoU & Dice & IoU & Dice & IoU \\ 
\midrule
CaraNet \cite{lou2022caranet}   & \textbf{81.39} & \textbf{68.82} & \textbf{81.74} & \textbf{69.25} & 85.01 & 74.06 \\
SegFormer \cite{xie2021segformer} & 78.83 & 65.34 & 80.86 & 68.01 & \textbf{85.65} & \textbf{75.01} \\
\bottomrule
\end{tabular}
\caption{Comparison of Segmentors for ITS}
\label{tab:ITS_backbone}
\end{table}

\begin{table*}[t!]
    \centering
    
    \begin{tabular}{p{33pt} p{33pt} p{33pt} p{33pt} p{33pt} p{33pt} p{33pt} p{33pt} p{33pt} p{33pt}}
    \toprule
        \multicolumn{4}{c}{}  & \multicolumn{2}{c}{CoNSep} & \multicolumn{2}{c}{MoNuSeg} & \multicolumn{2}{c}{TNBC} \\ \hline
        ITD & ITS & IDF & SCF & Dice & IoU & Dice & IoU & Dice & IoU \\
    \midrule
        \checkmark &         &       &         & 80.00 & 66.99 & 79.87 & 66.68 & 82.86 & 70.71 \\
                  & \checkmark &       &         & 79.86 & 65.54 & 79.38 & 65.97 & 80.66 & 66.18 \\
        \checkmark & \checkmark & \checkmark &         & 81.03 & 68.06 & 81.20 & 68.58 & 84.53 & 73.33 \\
        \checkmark & \checkmark &         & \checkmark & 80.36 & 67.24 & 80.47 & 67.70 & 82.98 & 70.87 \\
        \checkmark & \checkmark & \checkmark & \checkmark & \textbf{81.39} & \textbf{68.82} & \textbf{81.74} & \textbf{69.25} & \textbf{85.01} & \textbf{74.06} \\
    \bottomrule
    \end{tabular}
    \caption{Ablation Table}
    \label{tab:Ablation}
\end{table*}


\subsection*{Segmentation Backbones}
We utilized CaraNet as the backbone for training the Inference Time Segmentor (ITS) model. Additionally, we evaluated a transformer-based approach, SegFormer, for training ITS, as shown in Table \ref{tab:ITS_backbone}. The table compares the performance of BoxCell with CaraNet as the ITS model (BoxCell-CaraNet) and BoxCell with SegFormer as the ITS model (BoxCell-SegFormer). BoxCell-SegFormer demonstrated performance comparable to BoxCell-CaraNet. However, due to the limited dataset size, larger transformer-based models like SegFormer may not yield significant benefits, as evidenced in the results.

\subsection*{Discussion: Ablation and Error Analysis}
\subsubsection*{Ablation study}
We conduct an ablation study to evaluate the individual contributions of various components to the overall model performance, as summarized in Table~\ref{tab:Ablation}. Specifically, we examine the impact of the Intensity Distribution Factor (IDF) and the Spatial Constraining Factor (SCF), alongside the ITD and ITS predictions. Incorporating IDF into the ITD-ITS framework yields an improvement of 1–2  points, while SCF contributes a gain of 0.1–0.7 points—particularly beneficial in scenarios with high foreground–background contrast. The combination of all four components results in the best overall performance.

\subsubsection*{Solver Alternatives}
We evaluated BoxCell with multiple alternatives to Gurobi, including open-source ILP solvers (CBC, OR-Tools), approximate inference ($\alpha$-expansion), and a sparse graph formulation. Table \ref{tab:solver_comparison} report average time per image runtime, and Dice scores across datasets.
Across datasets, we find that open-source solvers (OR-Tools and CBC) and approximate methods ($\alpha$-expansion, sparse graph) achieve accuracy close to or within 1--2 Dice of the Gurobi baseline, while offering different trade-offs in runtime. The sparse graph formulation is especially effective, reducing runtime significantly while maintaining competitive Dice. These results demonstrate that BoxCell is not tied to a commercial solver: open-source or approximate alternatives can also be used.
\begin{table*}[h!]
\centering
\begin{tabular}{l|cc|cc|cc|c}
\hline
\multirow{2}{*}{Solver} & \multicolumn{2}{c|}{\textbf{CoNSep}} & \multicolumn{2}{c|}{\textbf{MoNuSeg}} & \multicolumn{2}{c|}{\textbf{TNBC}} & \multirow{2}{*}{Note} \\ 
 & Time/image (s) & Dice & Time/image (s) & Dice & Time/image (s) & Dice &  \\
\hline
CBC              & 204.25 & 78.99 & 219.00 & 75.74 & 190.00 & 81.58 & open-source \\
$\alpha$-expansion & 46.20 & 72.74 & 42.77 & 79.13 & 46.40 & 71.16 & approx. \\
Sparse graph     & 15.00 & 80.15 & 12.75 & 80.21 & 15.07 & \underline{85.02} & fast approx. \\
OR-Tools         & 87.30 & \underline{81.34} & 80.94 & \underline{80.60} & 85.59 & \textbf{85.03} & open-source \\
Gurobi           & 71.06 & \textbf{81.39} & 67.27 & \textbf{81.74} & 68.67 & 85.01& commercial \\
\hline
\end{tabular}
\caption{Comparison of solver performance across datasets. Gurobi achieves the best Dice scores consistently, while open-source alternatives (e.g., OR-Tools) provide competitive results.}
\label{tab:solver_comparison}
\end{table*}

\subsubsection*{Impact of Detector Threshold}
To analyze the impact of detector threshold, we varied the YOLOv8-m detection score threshold (0.1-0.9) and measured the resulting segmentation dice scores across datasets (see Table \ref{tab:ITD_threshold}). Segmentation quality is relatively stable across a wide range of detector thresholds, with only fluctuations in CoNSep at high threshold. Dice peaks at threshold of 0.3 for CoNSep, at 0.7 for MoNuSeg, and at 0.5 for TNBC. This indicates that BoxCell is not overly sensitive to the detector operating point: once bounding boxes are reasonably accurate, the ILP-based reconciliation mitigates detection errors.

\begin{table}[h]
\centering
\begin{tabular}{c|c|c|c}
\hline
Threshold & CoNSep (Dice) & MoNuSeg (Dice) & TNBC (Dice) \\
\hline
0.1 & 74.30 & 77.90 & 81.54 \\
0.2 & 76.30 & 77.90 & 81.54 \\
0.3 & 79.26 & 77.90 & 81.27 \\
0.4 & 73.20 & 77.89 & 81.86 \\
0.5 & 71.20 & 78.19 & 82.75 \\
0.6 & 70.30 & 78.94 & 82.08 \\
0.7 & 71.20 & 79.72 & 82.01 \\
0.8 & 68.90 & 79.73 & 80.11 \\
0.9 & 69.02 & 79.77 & 80.12 \\
\hline
\end{tabular}
\caption{Segmentation Dice scores across datasets at different YOLOv8-m detector thresholds.}
\label{tab:ITD_threshold}
\end{table}

\subsubsection*{Effect of Grid size $K$}
In our method, the grid size $K \times K$ is used to partition each image for local RGB-GMM modeling. In all our experiments we used $K=5$. To assess sensitivity, we varied $K$ in the range $\{3,5,10,15,20\}$. Table \ref{tab:K_sensitivity} provides dice score for varied value of K across the three datasets. As shown in table \ref{tab:K_sensitivity}, the performance remains stable across a wide range of $K$, with only marginal differences (within $\pm1$ Dice). This indicates that the method is not highly sensitive to the choice of grid size.

\begin{table}[h!]
\centering
\begin{tabular}{c|c|c|c}
\hline
$K$ & ConSep (Dice) & MoNuSeg (Dice) & TNBC (Dice) \\
\hline
3  & 81.38 & 81.07 & 85.30 \\
5  & 81.39 & 81.74 & 85.01 \\
10 & 81.36 & 80.34 & 84.62 \\
15 & 81.38 & 80.59 & 84.23 \\
20 & 81.23 & 80.76 & 84.52 \\
\hline
\end{tabular}
\caption{Sensitivity of BoxCell performance (Dice score) to grid size $K$.}
\label{tab:K_sensitivity}
\end{table}

\subsubsection*{Cross-domain Generalization and Stain Robustness}
We evaluate BoxCell against the strongest baseline in cross-domain settings (Table~\ref{tab:domain_generalization}), reporting train–test performance across datasets (C: CoNSeP, M: MoNuSeg, T: TNBC). BoxCell consistently outperforms SAM-BBTP++, demonstrating superior domain generalization. While performance decreases compared to within-domain results (81.39, 81.74, and 85.01 for CoNSeP, MoNuSeg, and TNBC, respectively), BoxCell still surpasses the closest baseline by effectively reconciling ITD and ITS. Training on one dataset and testing on another yields substantial Dice score gains (e.g., C $\rightarrow$ M: +22.63, C $\rightarrow$ T: +26.00, M $\rightarrow$ C: +24.98), further confirming better robustness to domain shifts.

\begin{table}[h!]
\centering
\begin{tabular}{lccc}
\toprule
\textbf{Method} & \textbf{Train $\rightarrow$ Test} & \textbf{Dice} \\
\midrule
SAM-BBTP++ & C $\rightarrow$ M & 35.24 \\
BoxCell (ours) & C $\rightarrow$ M & \textbf{57.87} \\
SAM-BBTP++ & C $\rightarrow$ T & 36.59 \\
BoxCell (ours) & C $\rightarrow$ T & \textbf{62.58} \\
\hline
SAM-BBTP++ & M $\rightarrow$ C & 32.13 \\
BoxCell (ours) & M $\rightarrow$ C & \textbf{57.11} \\
SAM-BBTP++ & M $\rightarrow$ T & 35.75 \\
BoxCell (ours) & M $\rightarrow$ T & \textbf{50.58} \\
\hline
SAM-BBTP++ & T $\rightarrow$ C & 26.42 \\
BoxCell (ours) & T $\rightarrow$ C & \textbf{49.67} \\
SAM-BBTP++ & T $\rightarrow$ M & 27.71 \\
BoxCell (ours) & T $\rightarrow$ M & \textbf{54.78} \\
\bottomrule
\end{tabular}
\caption{Cross-domain performance (train $\rightarrow$ test).}
\label{tab:domain_generalization}
\end{table}

\begin{figure*}[h!]
\centering
\begin{tikzpicture}[scale=1.0, transform shape, picture format/.style={inner sep=1pt}]

    \node at (1.5, 5) {\textbf{ConSeP}};
    \node at (4.25, 5) {\textbf{MoNuSeg}};
    \node at (7.0, 5) {\textbf{TNBC}};

    \node[rotate=90] at (-0.10,3.5) {\textbf{Original}};
    \node[rotate=90] at (-0.5,0.5) {\parbox{1cm}{\centering \textbf{Stain} \\ \textbf{Variation}}};

    \node[picture format] (A1) at (1.5,3.5) {\includegraphics[width=1.1in,height=1.0in]{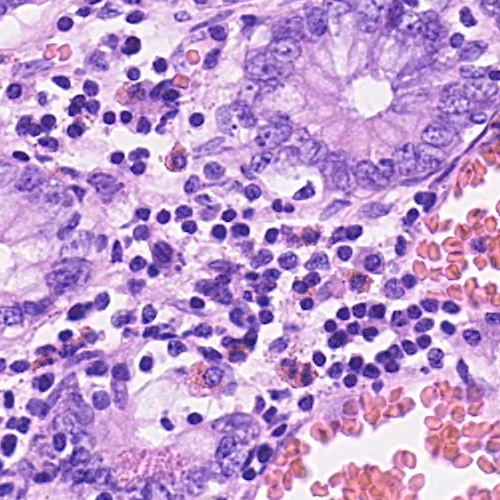}};
    \node[picture format,anchor=north] (B1) at (A1.south) {\includegraphics[width=1.1in,height=1.0in]{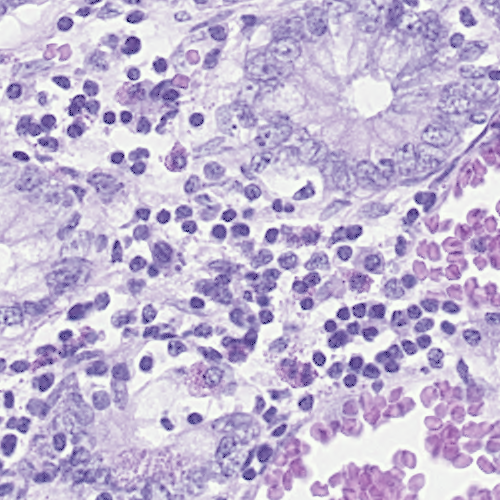}};

    \node[picture format,anchor=north west] (A2) at (A1.north east) {\includegraphics[width=1.1in,height=1.0in]{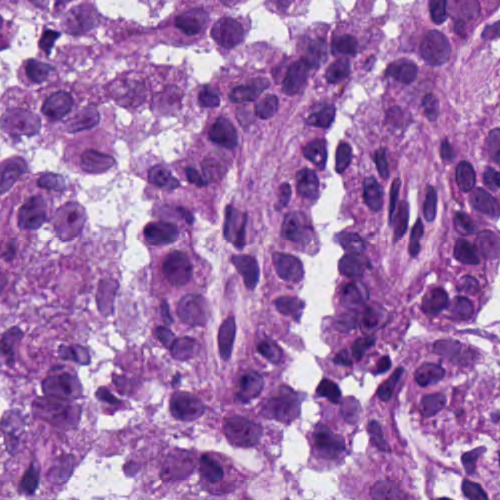}};
    \node[picture format,anchor=north] (B2) at (A2.south) {\includegraphics[width=1.1in,height=1.0in]{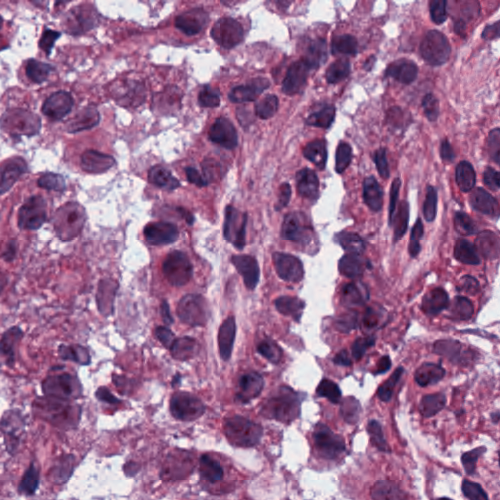}};

    \node[picture format,anchor=north west] (A3) at (A2.north east) {\includegraphics[width=1.1in,height=1.0in]{Images/stain_variation/tnbc_test_3_original.png}};
    \node[picture format,anchor=north] (B3) at (A3.south) {\includegraphics[width=1.1in,height=1.0in]{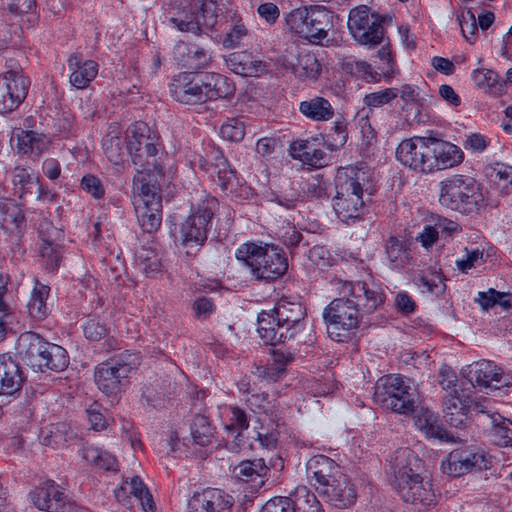}};

\end{tikzpicture}
\caption{Original and stain variation images across datasets.}
\label{fig:stain_variation}
\end{figure*}

Stain Robustness: To evaluate stain robustness, we generated stain-varied test sets for each dataset. Figure~\ref{fig:stain_variation} illustrates the original samples (row 1) and their stain-varied counterparts (row 2). Table~\ref{tab:stain_variation_results} shows that BoxCell demonstrates markedly greater robustness to stain variation compared to the baseline SAM-BBTP++. This indicates that BoxCell’s reconciliation of ITD and ITS effectively handles staining variability, making it a more reliable approach for real-world scenarios.

\begin{table}[h!]
\centering
\begin{tabular}{lcccc}
\hline
\textbf{Method} & & \textbf{CoNSep} & \textbf{MoNuSeg} & \textbf{TNBC} \\
\hline
SAM-BBTP++ & With Stain Variation & 63.20 & 65.59 & 62.72 \\
SAM-BBTP++ & No Stain Variation  & 74.64 & 73.58 & 79.64 \\
BoxCell &  With Stain Variation   & 73.80 &  78.20   &  80.75   \\
BoxCell & No Stain Variation    & 81.39 & 81.74 & 85.01 \\
\hline
\end{tabular}
\caption{Performance comparison across datasets under stain variation and no stain variation conditions.}
\label{tab:stain_variation_results}
\end{table}

\subsubsection*{Compute Capacity}
Table \ref{tab:time_comparision} provide parameters and time comparision for BoxCell and closest baseline. Both models have comparable parameter counts, ensuring similar capacity. The longer runtime of BoxCell arises mainly from the ILP optimization step, not from model size. While BoxCell is slower due to the ILP solver, it achieves substantially higher Dice scores across datasets. Another variant of BoxCell with a sparse solver (see row 2 in Table \ref{tab:time_comparision}), offers faster computation (15sec/image) while maintaining strong segmentation performance.

\begin{table}[h]
\centering
\begin{tabular}{lccccc}
\toprule
\textbf{Model} & \textbf{Parameters (M)} & \textbf{Time (sec/image)} & \textbf{ConSeP} & \textbf{MoNuSeg} & \textbf{TNBC} \\
\hline
SAM-BBTP & 15.2 & 0.2 & 74.64 & 73.58 & 79.64 \\
BoxCell - Sparse & 15.7 & 15 & 80.15 & 80.21 & \textbf{85.02} \\
BoxCell - Gurobi & 15.7 & 69.02 & \textbf{81.39} & \textbf{81.74} & 85.01 \\
\bottomrule
\end{tabular}
\caption{Compute Capacity comparison between BoxCell and SAM-BBTP}
\label{tab:time_comparision}
\end{table}

\subsubsection*{Error Analysis}

\begin{figure*}[t]
    \centering
    \includegraphics[scale = 0.5, , width = 16cm, height = 7cm]{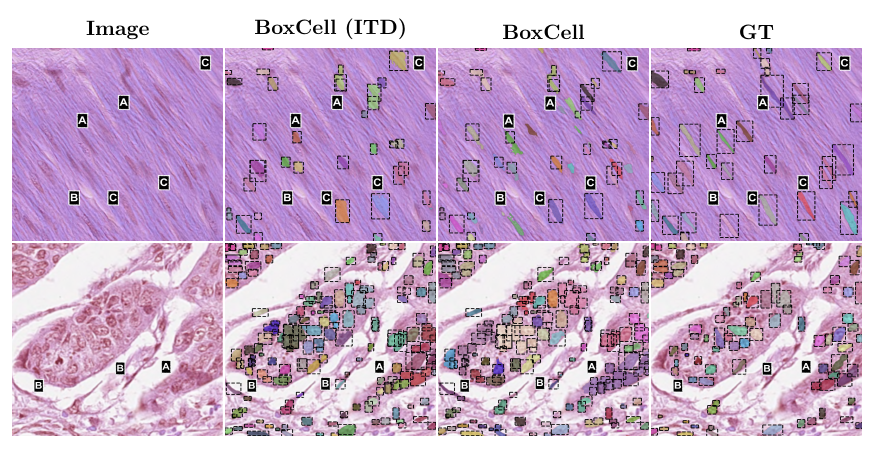}
    \caption{BoxCell with only ITD does not predict foreground outside box-prompts. BoxCell can do so, reducing the number of false negatives and improving the mask quality for wrongly sized boxes (\textit{A} and \textit{B}). BoxCell produces finer segmentation masks (\textit{C}). BoxCell performs less effectively for images with low contrast in f/g and b/g (\textit{A} row 2) where its capability to mitigate false positives is limited (\textit{B} in row 2).}
    \label{fig:err-analysis1}
\end{figure*}

On analyzing the masks qualitatively, we find that ITS benefits from a global perspective, leveraging the full image view for producing masks. As a result, ITS has a better shape understanding. Conversely, in ITD, SAM's inference is localized to the region defined by the box prompt. In scenarios where cell boundaries are ambiguous, ITS tends to overshoot the segmentation. In such cases, ITD often performs better, due to the localization provided by the box prompt. Another notable difference is the compounding of errors due to the pipelined nature of ITD. ITD is not able to recover from the mistakes of the object detector; if the box is a false positive, SAM generally outputs a false mask, and if a cell is missed by the detector, SAM can never output a mask for it. This is not an issue for ITS as it does not have a multi-step pipeline.  With the ILP, BoxCell mitigates issues of both component models. See Figure \ref{fig:err-analysis1} for an illustration. BoxCell demonstrates a reduced dependence on the size of bounding boxes -- it can make mask predictions outside the bounding boxes (\textit{A}, Row 1). It can generate segmentation masks even when the detection model predicts no bounding box, thus mitigating ITD's false negatives problem (\textit{B}, Row 1). Finally, it  yields qualitatively crisper boundaries, a characteristic only observed in BoxCell across all models ( \textit{C}, Row 1). 

Common failure modes for BoxCell include images with low contrast between the intensity of foreground and background pixels (\textit{A} in Fig \ref{fig:err-analysis1}, Row2). The human ability to detect such cells relies on the shape of these faintly different regions of intensities. The method fails to detect and mitigate false positives in such cases (\textit{B} in Fig \ref{fig:err-analysis1}, Row2). It tends to overlap instances of different cells due to no direct box supervision. It also tends to segment regions with intensity variation despite them not being cells. The detector's performance bottlenecks the performance of all detector-based models because of the high dependence on the box prompts. For the best-performing detection model, we still retain a lot of false positives that generate segmentation masks even when no actual cell is present. False negatives do not get segmented if the detector fails to detect them. Although SAM-ILP tries to mitigate these effects, they are still persistent in some cases.

\subsubsection*{Effect on Annotation Efficiency}


We observe that BoxCell substantially reduces the annotation time required by pathologists. To quantify this, we conducted an annotation-efficiency study with two pathologists, each annotating 25 randomly selected cells (total \(n=50\)). Manual polygon annotation, performed using LabelMe, required an average of 17.82 seconds per cell (95\% CI: 17.39--18.25, SD 1.5), including identifying cell boundaries, drawing polygons, and assigning class labels. In contrast, BoxCell requires only bounding box generation, taking 5.73 seconds per cell, followed by 4.20 seconds for verifying and refining the generated instance masks, resulting in a total of 9.92 seconds per cell (95\% CI: 8.93--10.91, SD 3.54). This represents a 7.9‑second reduction, corresponding to a 44.4\% improvement in annotation efficiency.

\section*{Conclusion}
\label{sec:conclusion}
We present BoxCell, the first approach to use SAM-based segmentation over histopathological images when only bounding box supervision is available. It computes two segmentation masks using SAM at train and test times; and reconciles them via a novel ILP that balances pixel likelihood and neighborhood objectives. Our experiments over three benchmark datasets show that our proposed approaches consistently beat the current weak supervision methods by up to 10 dice pts. Additionally, we compare with mask ensembling and refinement method and show the effectiveness of BoxCell. Our work opens up new possibilities for leveraging SAM in weak-supervision settings and using constrained optimization strategies to post-process segmentation masks.


\section*{Data availability}
The datasets used in this study are publicly available. The code for BoxCell implementation is available at https://github.com/dair-iitd/BoxCell. All experimental data and weights are available upon reasonable request to the corresponding author.

\bibliography{sample}








 








\section*{Acknowledgements}
We thank IIT Delhi HPC facility for computational resources.
This work is supported by Jai Gupta chair fellowship by IIT Delhi, Yardi School of AI, and Prime Minister Research fellowship (PMRF).

\section*{Author contributions statement}
Aayush: Writing—original draft, conceptualization, experiments, baseline implementation; Vaibhav: Baseline implementation, conceptualization, design; Prathosh A.P.: Writing—review and editing, conceptualization, design, supervision; Mausam: Writing—review and editing, conceptualization, design, methodology, resources, supervision.

\section*{Funding statement}
Project is supported by Indian Council of Medical Research (ICMR), Yardi School of AI, and Prime Minister Research fellowship (PMRF).


\section*{Competing interests}
The authors declare no competing interests.






\end{document}